\documentclass[letterpaper]{article} 
\usepackage{aaai2026}
\usepackage{times}  
\usepackage{helvet}  
\usepackage{courier}  
\usepackage[hyphens]{url}  
\usepackage{graphicx} 
\usepackage{natbib}  
\usepackage{caption} 
\usepackage[utf8]{inputenc} 
\usepackage[T1]{fontenc}    
\usepackage{url}            
\usepackage{booktabs}       
\usepackage{amsfonts}       
\usepackage{nicefrac}       
\usepackage{microtype}      
\usepackage{xcolor}         
\usepackage{amsmath}
\usepackage{bm}
\usepackage{subfig}
\usepackage{algorithm}
\usepackage{algorithmic}
\usepackage{multirow}
\usepackage{makecell}
\usepackage{adjustbox}  
\usepackage{tabularx} 
\usepackage{longtable}
\usepackage{threeparttable}

\usepackage{makecell}

\usepackage{newfloat}
\usepackage{listings}
\DeclareCaptionStyle{ruled}{labelfont=normalfont,labelsep=colon,strut=off} 
\floatstyle{ruled}
\newfloat{listing}{tb}{lst}{}
\floatname{listing}{Listing}
\title{Zero-Shot Human Mobility Forecasting via Large Language Model with Hierarchical Reasoning}

\author{
    Wenyao Li\textsuperscript{\rm 1,2},
    Ran Zhang\textsuperscript{\rm 1,2},
    Pengyang Wang\textsuperscript{\rm 3},
    Yuanchun Zhou\textsuperscript{\rm 2},
    Pengfei Wang\textsuperscript{\rm 2}
}
\affiliations{
    \textsuperscript{\rm 1}University of the Chinese Academy of Sciences\\
    \textsuperscript{\rm 2}Computer Network Information Center, Chinese Academy of Sciences\\
    \textsuperscript{\rm 3}Department of Computer and Information Science, University of Macau\\
    liwenyao25@mails.ucas.ac.cn
}

\begin{document}

\nocopyright

\maketitle

\begin{abstract}

Human mobility forecasting is crucial for enabling applications like transportation planning, urban management, and personalized recommendations.
Yet, existing methods often fail to generalize to unseen users or locations and struggle to capture dynamic intent due to limited labeled data and the inherent complexity of mobility patterns. 
To this end, we propose an innovative framework, \textbf{{Z}}ero-Shot \textbf{{H}}uman \textbf{{M}}obility \textbf{{F}}orecasting via Large Language Model with Hierarchical Reflection (ZHMF), combining a semantic-enhanced retrieval-reflection mechanism with a hierarchical language modeling system. 
The task is reformulated as a natural language question-answering paradigm.
By leveraging LLMs’ semantic understanding of user histories and context, our approach can effectively handle previously unseen prediction scenarios. 
Furthermore, we introduce a hierarchical reflection mechanism for iterative reasoning and refinement.
Specifically, we decompose forecasting process into an activity-level planner and a location-level selector, thus facilitating collaborative modeling of both long-term user intentions and short-term contextual preferences. 
Finally, extensive experimental results on classical human mobility datasets demonstrate that our approach significantly outperforms existing models.  
Comprehensive ablation studies reveal the importance of each module, while case studies offer further insights by showing that our approach effectively captures nuanced user intentions and dynamically adapts to diverse contextual scenarios.

\end{abstract}
\section{Introduction}
\label{sec:intro}

Accurate human mobility forecasting forms the foundation for a wide range of urban applications, including intelligent transportation planning, emergency response, and personalized location-based services. 
With recent advances in location-based and digital sensing services, unprecedented volumes of mobility trajectory data are now available, providing significant opportunities for predictive modeling. 
However, traditional methods face significant challenges, particularly in handling complex spatio-temporal dependencies, dynamic user preferences, as well as cold-start and zero-shot settings where users or contexts have not been observed during training. 
Overcoming these limitations requires the exploration of innovative technical paradigms.

Traditional human mobility modeling methods, such as Markov chain-based models \cite{Rendle2010FactorizingPM} and deep learning-based sequence models \cite{wang2017human,Kong2018HSTLSTMAH,Sun2020WhereTG,Luo2021STANSA}, have made strides in modeling user mobility. 
Recent developments in deep learning, graph neural networks(GNNs) have shown further promise. 
Sequence-based models like LSTM and Transformer \cite{Kong2018HSTLSTMAH,Sun2020WhereTG} capture temporal dynamics but often fail to model long-term dependencies effectively. 
GNNs, such as STHGCN \cite{Yan2023SpatioTemporalHL} and STP-UDGAT \cite{Lim2020STPUDGATSU}, leverage spatio-temporal graphs to enhance spatial representations while remaining limited in cold-start and zero-shot scenarios due to data sparsity. 
Meanwhile, hierarchical decision or planning models \cite{Zhang2023SpatialTemporalII,Xiao2024HierarchicalRL} are capable of capturing multi-layered decision-making processes, such as high-level goal planning and low-level action selection. However, they still struggle to efficiently adapt to sparse user interactions and to provide interpretable predictions in unfamiliar contexts.


Recent advancements in Large Language Models (LLMs) reveal significant breakthroughs in semantic understanding, offering the potential to better comprehend latent spatio-temporal interactions in mobility trajectories~\cite{Wang2023WhereWI,Wang2024SeCorAS,Feng2024WhereTM,Wu2024MAS4POIAM}. 
Leveraging the powerful semantic reasoning capabilities of LLMs can help address cold-start and zero-shot challenges, making user mobility modeling more robust across diverse and unseen contexts. 
Nevertheless, LLMs sometimes generate hallucinated information or lack grounding in real-world scenarios, which can limit their effectiveness for mobility forecasting. 
Simultaneously, hierarchical reasoning frameworks excel at decomposing complex decision-making into multiple levels, enabling high-level goal planning and low-level action selection.
Thus, integrating LLM-driven semantic reasoning with a structured hierarchical decision-making framework represents a promising and largely unexplored direction for advancing zero-shot human mobility forecasting.

\begin{figure}
\centering
\includegraphics[width=1\columnwidth]{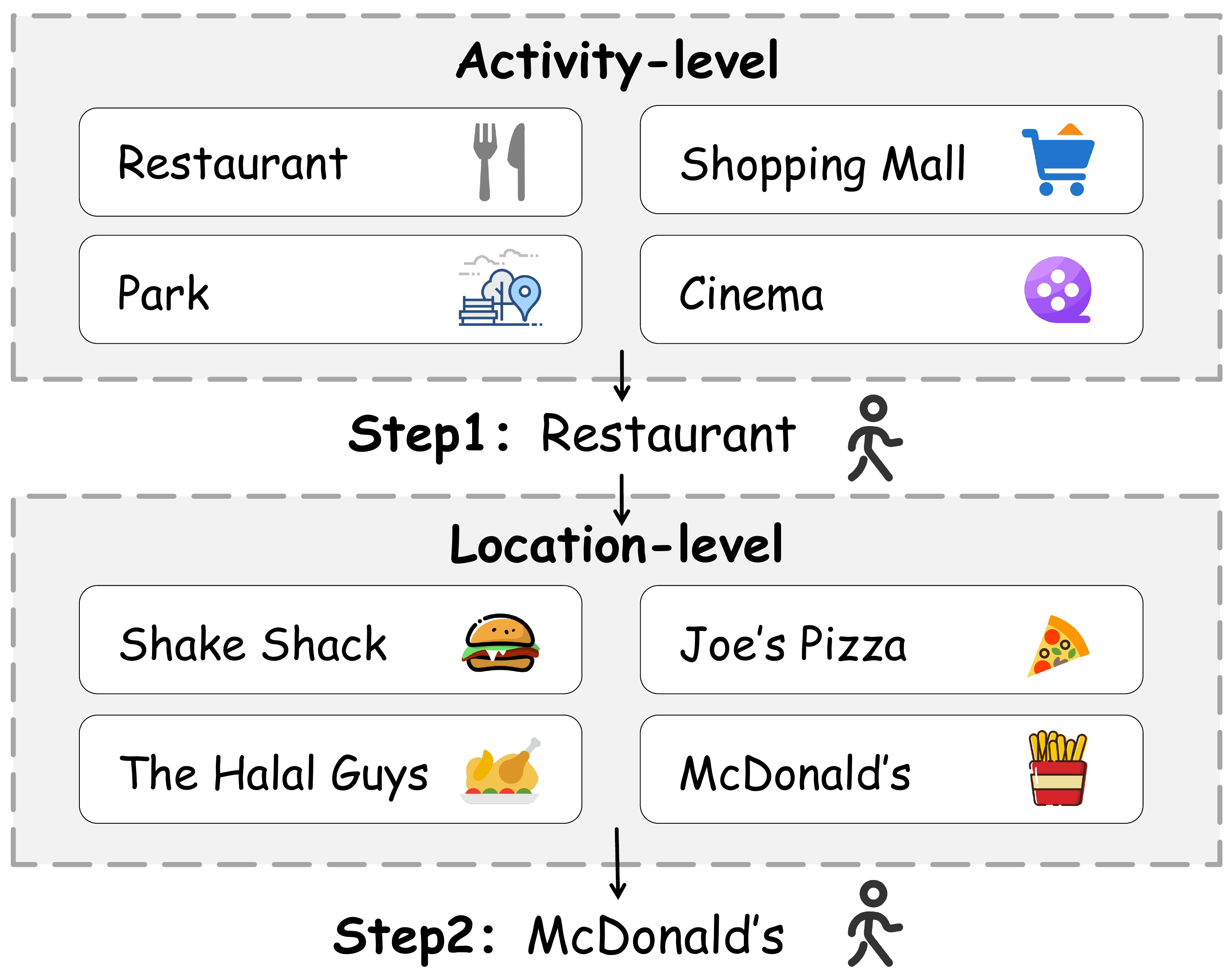}
\caption{Toy example of hierarchical mobility framework.}
\label{fig:toy_example}
\end{figure}


Specifically, our approach leverages LLMs to model spatio-temporal dependencies and dynamic user preferences by reformulating the trajectory prediction task as a natural language question-answering problem, enabling flexible interpretation of historical trajectories and contextual information. 
Meanwhile, it introduces a hierarchical reflective mechanism that adaptively combines short-term trajectory data with structured long-term memory to enhance contextual awareness. 
The systematic modeling of activity-level and location-level reasoning further enables robust generalization in complex, real-world scenarios. 
Notably, all LLM components in our system are employed as frozen backbones via prompt engineering only, and no fine-tuning or gradient-based updates are conducted on the language models themselves. 
For example, as Figure ~\ref{fig:toy_example} shows, a person first decides to eat at a restaurant and then chooses a specific location like McDonald's.   

The main contributions are summarized as follows: 
 
\noindent$\bullet$ 
Our method introduces a novel integration of LLMs' semantic understanding with a hierarchical reasoning framework for human mobility forecasting, enabling more context-aware predictions.

\noindent$\bullet$ The proposed dynamic prompt engineering reformulates mobility intelligence tasks into a unified natural language reasoning paradigm, employing a two-stage inference process that first determines the user’s activity type and then predicts the specific location.

\noindent$\bullet$ We propose a reflective retrieval mechanism that integrates short-term trajectory records with structured long-term memory. 
By leveraging collective mobility experiences, this approach achieves accurate human mobility forecasting, even under cold-start and zero-shot scenarios.

\noindent$\bullet$ Evaluations on Foursquare and Gowalla datasets show our framework outperforms state-of-the-art models, with ablation studies validating key modules' effectiveness.
\section{Related Work}
\label{sec:relat}

Recent years have witnessed advances in human mobility forecasting. 
We provide a brief overview of key developments in this field. 
Details are available in supplementary materials. 


\noindent\textbf{Traditional Mobility Modeling} methods frequently rely on sequence modeling techniques \cite{10.5555/2540128.2540504} first proposed the next POI recommendation task through local region constraints in FPMC \cite{Rendle2010FactorizingPM}. 
With the rise of deep learning, models like STAN \cite{Luo2021STANSA} employed multimodal embeddings and attention mechanisms to model spatio-temporal correlations. 
Despite these advances, traditional methods struggle with scalability, cold-start and zero-shot scenarios, and adapting to dynamic user preferences. 
These limitations have driven researchers toward graph neural networks (GNNs), with works such as STHGCN \cite{Yan2023SpatioTemporalHL} addressing user behavior modeling and improving generalization capabilities.


\noindent\textbf{Hierarchical Modeling} have been widely explored for human mobility analysis and recommendation systems by decomposing complex decision-making processes into multi-level subtasks, which facilitates user interest modeling and structured decision making.
For example, STI-HRL \cite{Zhang2023SpatialTemporalII} introduced hierarchical decoupling to the human mobility prediction process, while HRL-PRP \cite{Xiao2024HierarchicalRL} utilized high-level and low-level task decomposition during user profile optimization in the preprocessing stage.
These advancements demonstrate the potential of hierarchical modeling for capturing dynamic user preferences and improving recommendation accuracy in complex scenarios.

\noindent\textbf{Large Language Models} have gained broad attention in mobility modeling and POI recommendation tasks due to their exceptional natural language processing capabilities and ability to handle heterogeneous data \cite{Minaee2024LargeLM}. 
By transforming complex recommendation problems into intuitive question-answering paradigms, LLMs integrate multi-dimensional information while leveraging built-in common-sense knowledge to enhance user behavior modeling. 
For human mobility forecasting, Wang et al. \cite{Wang2023WhereWI} initiated the application of contextual learning, while LLMmove \cite{Feng2024WhereTM} generates next POI recommendations without task-specific training, and LLM4POI \cite{Li2024LargeLM} fine-tunes LLMs with trajectory similarity. 
Emerging trends focus on zero-shot human mobility forecasting, which reduces reliance on large-scale labeled data. 
However, current LLM-based approaches still face challenges in fully capturing geographical contexts and handling complex decision-making processes.

\begin{figure*}[htbp]
\centering
\includegraphics[width=0.95\linewidth]
{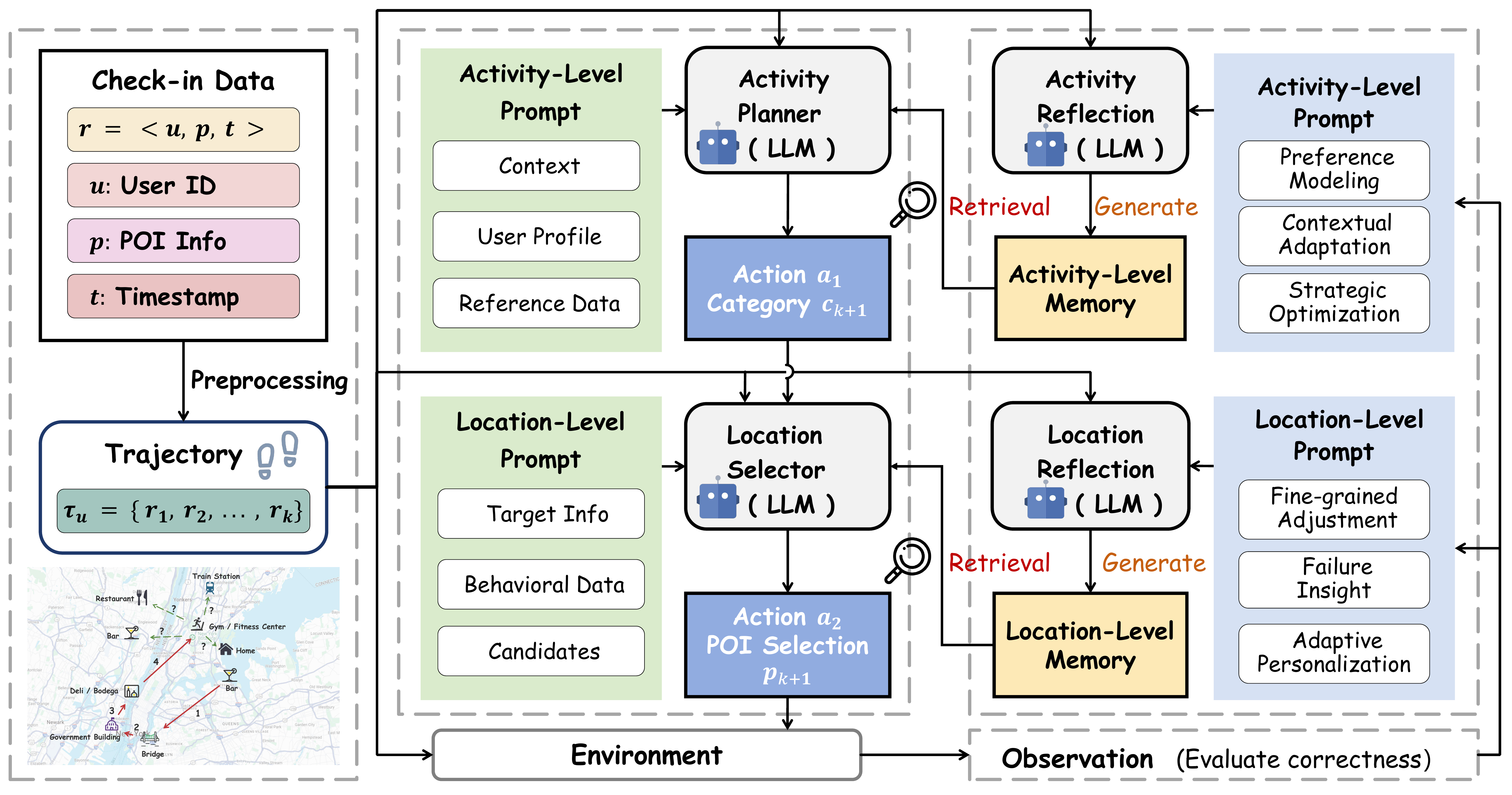}
\caption{The overall of our ZHMF framework.}
\label{fig:framework}
\end{figure*}
\section{Problem Statement}  
\label{sec:probl}

\noindent\textbf{POI and User Representation.} Let $\mathcal{U} = \{u_1, u_2, \ldots, u_N\}$ denote the set of users and $\mathcal{P} = \{p_1, p_2, \ldots, p_M\}$ the set of POIs, where $N$ and $M$ are the total numbers of users and POIs, respectively. Each POI $p \in \mathcal{P}$ is represented as $p = \langle \mathit{id}, \mathit{cat}, \mathit{lat}, \mathit{lon} \rangle$, where $\mathit{id}$ is a unique identifier, $\mathit{cat} \in \mathcal{C}$ is the category label (e.g., Restaurant, Bookstore), and $\mathit{lat}, \mathit{lon}$ specify the geographical coordinates.

\noindent\textbf{Check-in Records and Trajectories.} A check-in record is defined as a tuple $r = \langle u, p, t \rangle$, indicating that user $u \in \mathcal{U}$ visited POI $p \in \mathcal{P}$ at timestamp $t$. A trajectory $\tau_u = \{r_1, r_2, \ldots, r_k\}$ represents a sequence of chronologically ordered check-ins by user $u$ within a specific time interval, where $r_i = \langle u, p_i, t_i \rangle$ and $t_1 < t_2 < \cdots < t_k$. For simplicity, we can also denote a trajectory as a sequence of POIs $\tau_u = \{p_1, p_2, \ldots, p_k\}$ when the user and temporal information are clear from context.  

\noindent\textbf{Next POI Recommendation Task.} Given a user $u$'s historical trajectory $\tau_u = \{p_1, p_2, \ldots, p_k\}$ and the current contextual information (including timestamp, location, and user profile), the next POI recommendation task aims to predict the subsequent POI $p_{k+1}$ that user $u$ is most likely to visit.



\section{Method}
\label{sec:method}

This study proposes a hierarchical reasoning framework based on LLMs to address key challenges in human mobility modeling and location-aware recommendation. As shown in Figure~\ref{fig:framework}, the proposed method consists of three core modules: Prompt-driven Semantic Understanding Module, Hierarchical Reasoning and Planning Module, and Retrieval-augmented Reflective Memory. 
The LLM backbone remains frozen with no parameter fine-tuning; all adaptation occurs through prompt engineering and external reflective memory updates.
The Semantic Understanding and Task Transformation Module transforms users' historical trajectory data and contextual information into structured natural language descriptions using prompt templates. The hierarchical reasoning framework decomposes the decision process into two levels: the activity-level reasoning identifies target activity categories, while the location-level reasoning selects specific POI recommendations based on the activity-level output. The Reflective Memory Retrieval Module, with short-term and long-term memory components, stores user behavior patterns and incorporates a reflection mechanism to dynamically refine the recommendation reasoning process.

\begin{figure}[htbp]
    \centering
    \includegraphics[width=1\linewidth]{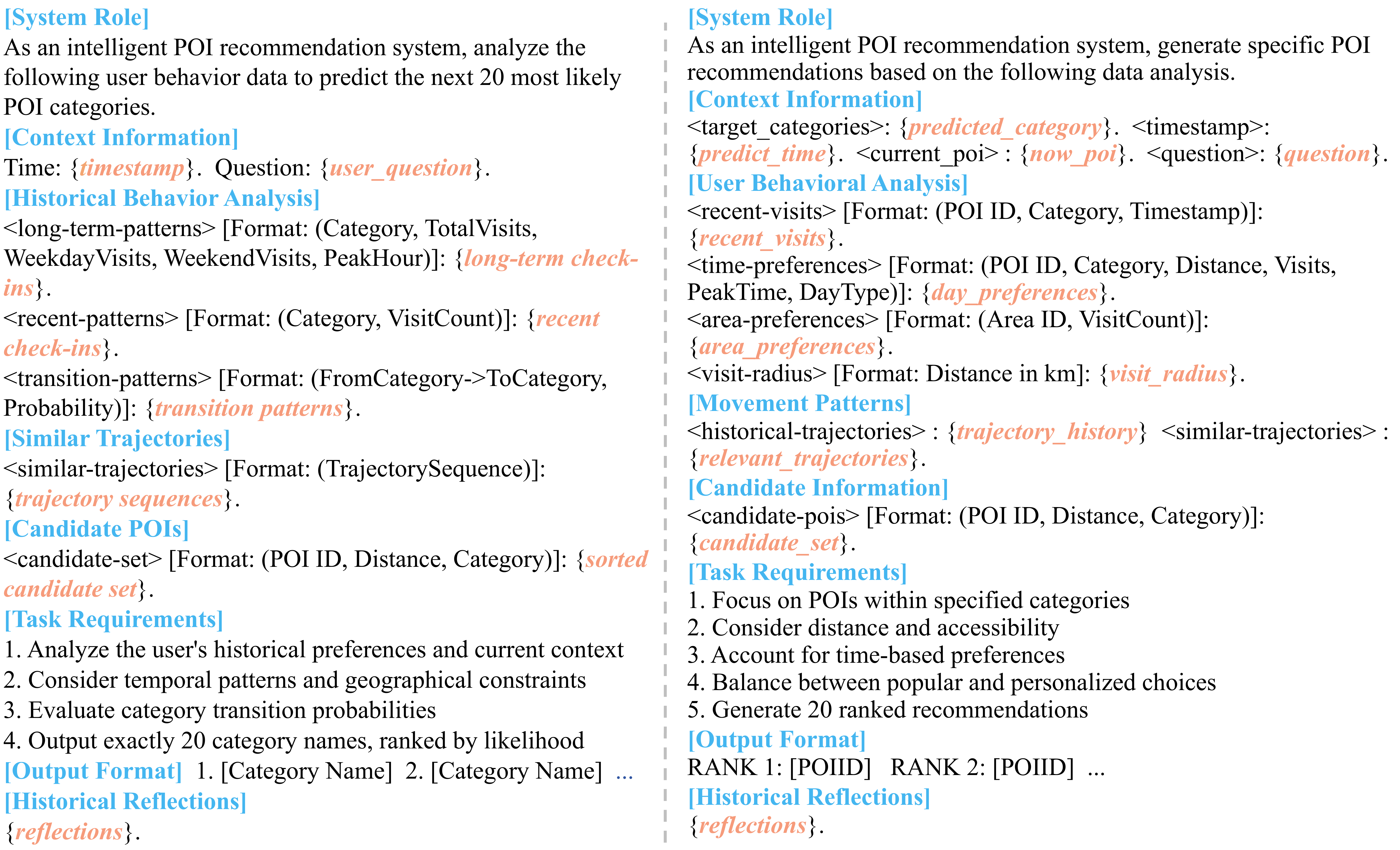}
    \caption{Reasoning prompt template: the left part shows the activity-level prompt template, and the right part shows the location-level prompt template.}
    \label{fig:prompt}
\end{figure}

\noindent\textbf{Prompt-driven Semantic Understanding.}
The semantic understanding module leverages LLMs to reframe mobility modeling as natural language processing problems. This module has three main components: task transformation, prompt template design, and output parsing. As shown in Figure~\ref{fig:framework}, the task transformation module converts users' historical check-in sequences and contextual information into natural language descriptions, retaining key data like user IDs, timestamps, and locations. The prompt template design, illustrated in Figure~\ref{fig:prompt}, creates structured templates that guide the model in generating accurate recommendations. These prompts combine task instructions, user profiles, historical behaviors, and environmental contexts. The reflection enhancement component integrates insights from past experiences to refine the prompts. Finally, the output parsing module converts the natural language recommendations into structured forms, extracting categories and specific POI lists for downstream processing.


\noindent\textbf{Hierarchical Reasoning and Planning Module.}
The proposed hierarchical reasoning framework enables flexible and robust human mobility intelligence by decomposing the next POI prediction task into structured multi-level reasoning steps. As shown in Figure~\ref{fig:framework}, this method tackles mobility-related decision processes by dividing decision-making into two levels: activity-level  goal generation and location-level specific recommendations. The framework integrates modular language feedback, utilizing the semantic understanding capabilities of LLMs to dynamically adapt to user preferences and environmental changes. Unlike traditional reinforcement learning methods, it eliminates the dependency on frequent parameter updates while enhancing flexibility and interpretability. The activity-level reasoning module focuses on long-term user preferences by transforming historical behaviors and spatio-temporal contexts into natural language descriptions. It generates broad activity categories, such as "restaurant" or "café," which are passed to the location-level reasoning module. The location-level module refines these categories into specific POI recommendations by combining short-term user behaviors, candidate POI attributes, and contextual spatio-temporal information. Modular language feedback plays a key role throughout the framework, converting behavioral data and environmental information into structured natural language representations to facilitate dynamic adjustment. Specifically, the hierarchical approach consists of the following two main steps:

\noindent(1) Activity-level Category Prediction.  
The activity-level reasoning function predicts the next POI category $c_{k+1}$ by considering the user’s historical trajectory $\tau_u$ and contextual information $\mathcal{C}_t$ (including timestamp, spatial context, and user preferences):  
\begin{equation}
c_{k+1} = \pi_A(\tau_u, \mathcal{C}_t, \Phi_A),
\end{equation}
where $\tau_u$ denotes the user’s historical POI sequence, $\mathcal{C}_t$ is the current context (e.g., time, recent location, user profile), $\pi_A$ is the activity-level reasoning function, and $\Phi_A$ is a structured prompt template encoding behavioral, contextual, and reflective information for the LLM.

\noindent(2) Location-level POI Recommendation.  
Given the predicted category, the location-level policy selects the specific POI candidate:
\begin{equation}
p_{k+1} = \pi_L(c_{k+1}, \tau_u, \mathcal{C}_t, \Phi_L),
\end{equation}
where $\pi_L$ is the location-level reasoning function and $\Phi_L$ is a prompt including short-term visits, movement patterns, candidate POIs, and reflections.

The two-level prompt templates are further structured as:
\begin{equation}
\begin{split}
\Phi_A = \langle \mathcal{B}_L, \mathcal{B}_R, \mathcal{T}, \mathcal{S}, \mathcal{R}_A^n \rangle, \\
\Phi_L = \langle \mathcal{V}_R, \mathcal{V}_T, \mathcal{V}_A, \mathcal{M}, \mathcal{P}_C, \mathcal{R}_L^n \rangle
\end{split}
\end{equation}
where $\mathcal{B}_L$ and $\mathcal{B}_R$ represent long-term and recent category patterns, $\mathcal{T}$ is the transition statistic between categories, $\mathcal{S}$ denotes similar trajectory samples, and $\mathcal{R}_A^n$ is the top-$n$ retrieved activity-level (category) reflections from the memory. $\mathcal{V}_R$ is recent POI visits, $\mathcal{V}_T$ represents time-based POI preferences, $\mathcal{V}_A$ is area-level spatial preference, $\mathcal{M}$ captures movement history, $\mathcal{P}_C$ is the set of candidate POIs in $c_{k+1}$, and $\mathcal{R}_L^n$ is the top-$n$ retrieved location-level (POI) reflections from the memory. The optimal value of $n$ is determined through ablation studies presented in Section~\ref{sec:hyperparameter}.

\begin{figure}[htbp]
\centering
\includegraphics[width=1\linewidth]{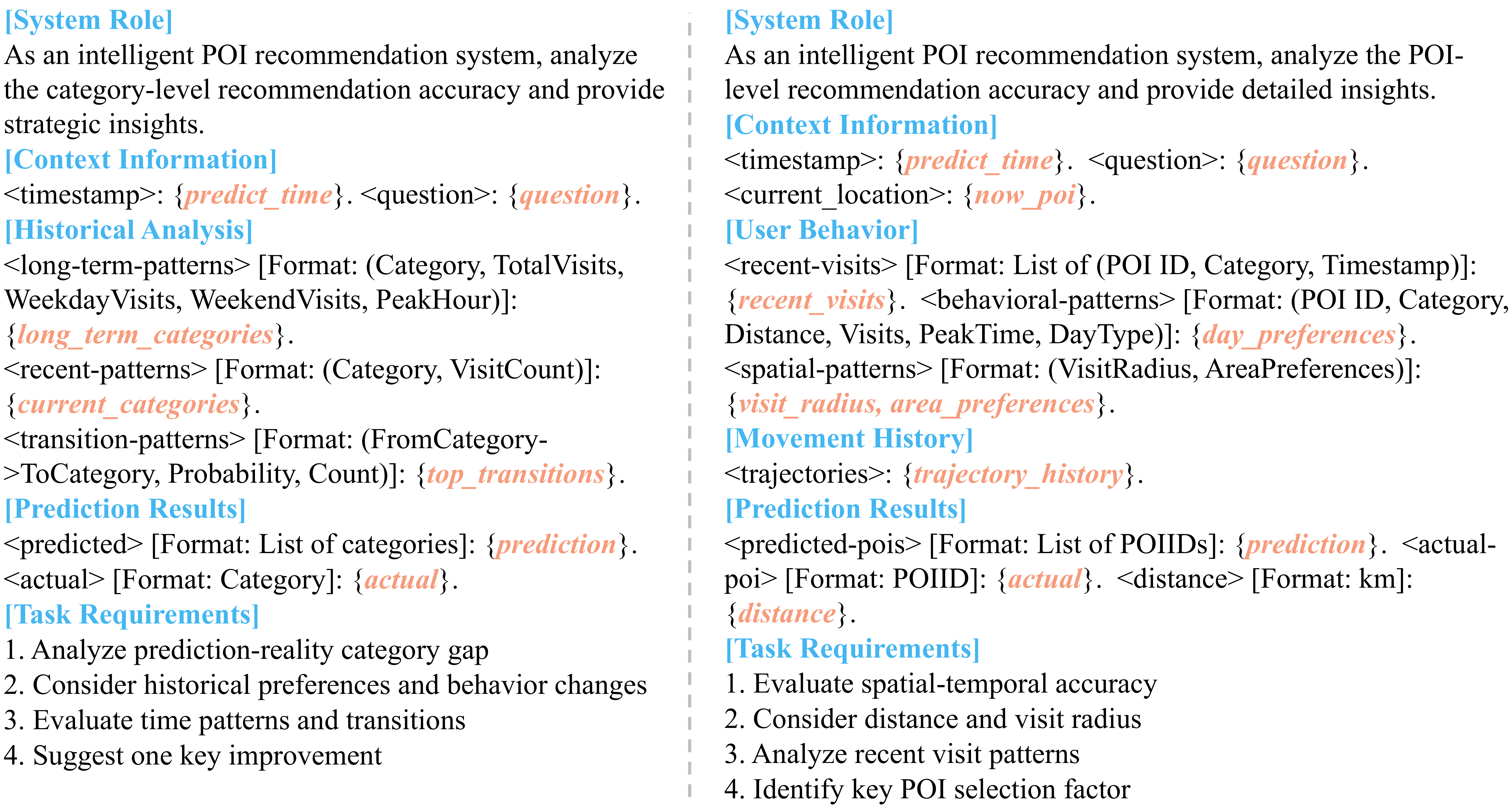}
\caption{Reflection prompt template: the left part shows the activity-level prompt template, and the right part shows the location-level prompt template.}
\label{fig:reflection prompt}
\end{figure}

\noindent\textbf{Retrieval-augmented Reflective Memory.}
The Reflective Retrieval Memory Module is a core innovation of this framework, providing knowledge support by integrating short-term trajectory storage and long-term hierarchical reflective memory accumulation. As shown in Figure~\ref{fig:reflection prompt}, the memory system categorizes reflections into activity-level  goal reflections and location-level action reflections. This hierarchical design separates global goal and local action reflections, enhancing efficiency and precision in strategy adjustments~\cite{Xiao2024HierarchicalRL,Zhang2023SpatialTemporalII}. The Short-term Memory Module stores user trajectory data such as visit timestamps, POI categories, and locations, constructing a fine-grained behavioral context for immediate recommendations. The Long-term Memory Module accumulates decision experiences at two levels: activity-level  memory stores category-level reflections, while location-level memory focuses on specific POI recommendations. Reflections are generated via a structured language mechanism, using templates tailored to activity-level  strategies and location-level actions, as depicted in Figure~\ref{fig:reflection prompt}. When a recommendation fails, the system records contextual data, analyzes reasons for failure with LLMs, and produces actionable reflection content. Through semantic matching, the memory retrieval mechanism activates relevant knowledge based on current short-term contexts, efficiently integrating past insights into decision prompts.

Through semantic matching, the retrieval module encodes both the current query and historical memory items into dense vectors, and computes the cosine similarity to identify the most relevant entries:
\begin{equation}
\textrm{sim}(e_q, e_i) = \frac{e_q \cdot e_i}{\|e_q\| \|e_i\|}.
\end{equation}
The top-$n$ most similar items are then retrieved and denoted as $\mathcal{R}_A^n$ and $\mathcal{R}_L^n$ for activity and location levels respectively, which are incorporated into the corresponding prompts. This retrieval mechanism is applied for both reflective memory and similar trajectory search.

For continual reasoning refinement, the activity-level reflection function $\rho_A$ updates category-level memory by comparing predicted and actual user behavior:
\begin{equation}
\rho_A(c_{k+1}^{(pred)}, c_{k+1}^{(actual)}, \tau_u, \mathcal{C}_t) \rightarrow \mathcal{R}_A
\end{equation}
where $\rho_A$ updates activity-level memory based on prediction outcomes and context.

Similarly, the location-level reflection $\rho_L$ focuses on specific POI recommendations:
\begin{equation}
\rho_L(p_{k+1}^{(pred)}, p_{k+1}^{(actual)}, c_{k+1}, \tau_u, \mathcal{C}_t) \rightarrow \mathcal{R}_L
\end{equation}
where $\rho_A$ and $\rho_L$ are functions that generate new reflection feedback for activity-level and location-level memory, respectively, based on prediction outcomes and context.


\section{Experiment}
\label{sec:experi}

\subsection{Experimental Setup}
\label{sec:experiment_setup}

\textbf{Datasets.} 
We conduct experiments on three widely used location-based social network datasets: Foursquare-NYC, Foursquare-TKY, and Gowalla-CA. 
We follow the preprocessing procedure of Yan et al.~\cite{Yan2023SpatioTemporalHL}, filtering out POIs and users with fewer than 10 check-ins and constructing trajectory sequences based on 24-hour time intervals. 
For data partitioning, we chronologically split the check-in records into 80\% for training, 10\% for validation, and 10\% for testing, ensuring that validation and test sets only contain users and POIs present in the training set. 
Detailed dataset statistics and preprocessing information are provided in supplementary materials.  
For LLM-based evaluation, following LLMMove~\cite{Feng2024WhereTM}, all candidate POIs are presented to the model in ascending order of distance from the user's current location when constructing prompts, which ensures reliable and fair evaluation.

\noindent\textbf{Baselines.} 
We compare our method with several state-of-the-art baselines from different categories. 
For traditional recommendation systems, we include MF \cite{Koren2009MatrixFT} and FPMC \cite{Rendle2010FactorizingPM}. 
For sequential modeling approaches, we select LSTM \cite{Hochreiter1997LongSM}, PRME \cite{Feng2015PersonalizedRM}, and ST-RNN \cite{Liu2016PredictingTN}. 
For spatiotemporal methods, we include STGN \cite{Zhao2019WhereTG}, STGCN \cite{Zhao2019WhereTG}, PLSPL \cite{Wu2019LongAS}, and STAN \cite{Luo2021STANSA}. 
For LLM-based approaches, we compare with LLMMob \cite{Wang2023WhereWI} and LLMMove \cite{Feng2024WhereTM}. 
Recent works such as Mobility-LLM~\cite{Gong2024MobilityLLMLV} and LLM4POI~\cite{Li2024LargeLM} involve task-specific fine-tuning on mobility datasets. Since our focus is on fully zero-shot LLM-based methods, we do not include direct comparisons with these approaches.

\noindent\textbf{Evaluation Metrics.}
To comprehensively evaluate the performance of our proposed model, we adopt the following commonly used recommendation system evaluation metrics: Acc@1, Acc@5, Acc@10, Acc@20, and MRR (Mean Reciprocal Rank).

\noindent\textbf{Implementation Details.}
Our model is implemented in PyTorch (v2.5.1) and runs on Ubuntu 18.04. We use Meta-Llama-3-8B-Instruct as the main language backbone and BAAI/bge-small-en-v1.5 for sentence embeddings. All experiments are conducted on workstations equipped with NVIDIA A100 (40GB) GPUs.
More details can be found in supplementary materials.

\begin{table*}[htbp]
    \centering  
    \begin{threeparttable}
    \small
    \setlength{\tabcolsep}{1.2pt}
    \caption{Performance comparison in Acc@k and MRR on three datasets (higher is better). The best scores are highlighted in \textbf{bold}, and the second-best are \underline{underlined}.}  
    \label{tab:performance-comparison}  
    \begin{tabularx}{\textwidth}{l*{5}{>{\centering\arraybackslash}X}|*{5}{>{\centering\arraybackslash}X}|*{5}{>{\centering\arraybackslash}X}}
        \toprule
        \multirow{2}{*}{Model} & \multicolumn{5}{c|}{NYC} & \multicolumn{5}{c|}{TKY} & \multicolumn{5}{c}{CA} \\
        \cmidrule(lr){2-6} \cmidrule(lr){7-11} \cmidrule(lr){12-16}
        & Acc@1 & Acc@5 & Acc@10 & Acc@20 & MRR 
        & Acc@1 & Acc@5 & Acc@10 & Acc@20 & MRR 
        & Acc@1 & Acc@5 & Acc@10 & Acc@20 & MRR \\
        \midrule
        MF & 0.0368 & 0.0961 & 0.1522 & 0.2375 & 0.0672 & 0.0241 & 0.0701 & 0.1267 & 0.1845 & 0.0461 & 0.0110 & 0.0442 & 0.0723 & 0.1190 & 0.0342 \\
        FPMC & 0.1003 & 0.2126 & 0.2970 & 0.3323 & 0.1701 & 0.0814 & 0.2045 & 0.2746 & 0.3450 & 0.1344 & 0.0383 & 0.0702 & 0.1159 & 0.1682 & 0.0911 \\
        LSTM & 0.1305 & 0.2719 & 0.3283 & 0.3568 & 0.1857 & 0.1335 & 0.2728 & 0.3277 & 0.3598 & 0.1834 & 0.0665 & 0.1306 & 0.1784 & 0.2211 & 0.1201 \\
        PRME & 0.1159 & 0.2236 & 0.3105 & 0.3643 & 0.1712 & 0.1052 & 0.2278 & 0.2944 & 0.3560 & 0.1786 & 0.0521 & 0.1034 & 0.1425 & 0.1954 & 0.1002 \\
        ST\mbox{-}RNN & 0.1483 & 0.2923 & 0.3622 & 0.4502 & 0.2198 & 0.1409 & 0.3022 & 0.3577 & 0.4753 & 0.2212 & 0.0799 & 0.1423 & 0.1940 & 0.2477 & 0.1429 \\
        STGN & 0.1716 & 0.3381 & 0.4122 & 0.5017 & 0.2598 & 0.1683 & 0.3391 & 0.3848 & 0.4514 & 0.2422 & 0.0819 & 0.1842 & 0.2579 & 0.3095 & 0.1675 \\
        STGCN & 0.1799 & 0.3425 & 0.4279 & 0.5214 & 0.2788 & 0.1716 & 0.3453 & 0.3927 & 0.4763 & 0.2504 & 0.0961 & 0.2097 & 0.2613 & 0.3245 & 0.1712 \\
        PLSPL & 0.1917 & 0.3678 & 0.4523 & 0.5370 & 0.2806 & 0.1889 & 0.3523 & 0.4150 & 0.4850 & 0.2542 & 0.1072 & 0.2278 & 0.2995 & 0.3401 & 0.1847 \\
        STAN
        & \underline{0.2231} & \underline{0.4532} & \underline{0.5734} & \underline{0.6328} & \underline{0.3253} 
        & \underline{0.1963} & \underline{0.3798} & \underline{0.4464} & \underline{0.5119} & \underline{0.2852} 
        & \underline{0.1104} & \underline{0.2348} & \underline{0.3018} & \underline{0.3502} & \underline{0.1869} \\
        LLMMob\tnote{*} & 0.0661 & 0.1492 & 0.2145 & 0.3534 & 0.1137 & 0.0144 & 0.0434 & 0.0520 & 0.0574 & 0.0371 & 0.0192 & 0.0419 & 0.0476 & 0.0516 & 0.0368 \\
        LLMMove\tnote{*} & 0.0772 & 0.1284 & 0.1678 & 0.2353 & 0.1045 & 0.0574 & 0.1005 & 0.1368 & 0.1901 & 0.0805 & 0.0590 & 0.1086 & 0.1237 & 0.1450 & 0.0809 \\
        \midrule  
        \textbf{ZHMF} 
        & \textbf{0.3437} & \textbf{0.6325} & \textbf{0.6904} & \textbf{0.7008} & \textbf{0.4596} 
        & \textbf{0.3142} & \textbf{0.5375} & \textbf{0.5872} & \textbf{0.6009} & \textbf{0.4062} 
        & \textbf{0.2734} & \textbf{0.4824} & \textbf{0.5460} & \textbf{0.5691} & \textbf{0.3581} \\
        \bottomrule
    \end{tabularx}
    \begin{tablenotes}
      \scriptsize
      \item[*] LLMMob~\cite{Wang2023WhereWI} and LLMMove~\cite{Feng2024WhereTM} results are reproduced with Llama3 under the same settings as ZHMF.
    \end{tablenotes}  
    \end{threeparttable}
\end{table*}

\subsection{Overall Performance}
Table~\ref{tab:performance-comparison} shows our ZHMF model consistently outperforms all baseline methods across three datasets. On NYC, our model achieves an Acc@1 of 34.37\% and MRR of 45.96\%, significantly improving over the strongest baseline STAN \cite{Luo2021STANSA}. For TKY, we maintain superior performance with 31.42\% Acc@1, highlighting the advantage of our hierarchical approach in understanding complex urban mobility patterns. 
Notably, the improvements of ZHMF are not limited to a single metric, but are observed across Acc@1, Acc@5, Acc@10, Acc@20, and MRR, indicating the model’s robust capability in both ranking the most relevant POIs and maintaining overall recommendation quality.
Even on the challenging Gowalla-CA dataset with sparser check-in data, ZHMF demonstrates robust generalization with an Acc@1 of 27.34\%. Furthermore, the performance gap between ZHMF and the baselines remains significant across different datasets, which suggests that the proposed method consistently adapts well to varying levels of data sparsity and mobility complexity.
Notably, our approach consistently achieves better performance than LLM-based methods such as LLMMob \cite{Wang2023WhereWI} and LLMMove \cite{Feng2024WhereTM}, as well as other existing sequential and graph-based methods.

\subsection{Zero-shot Forecasting}

\begin{figure}[htbp]
    \centering
    \includegraphics[width=1\linewidth]{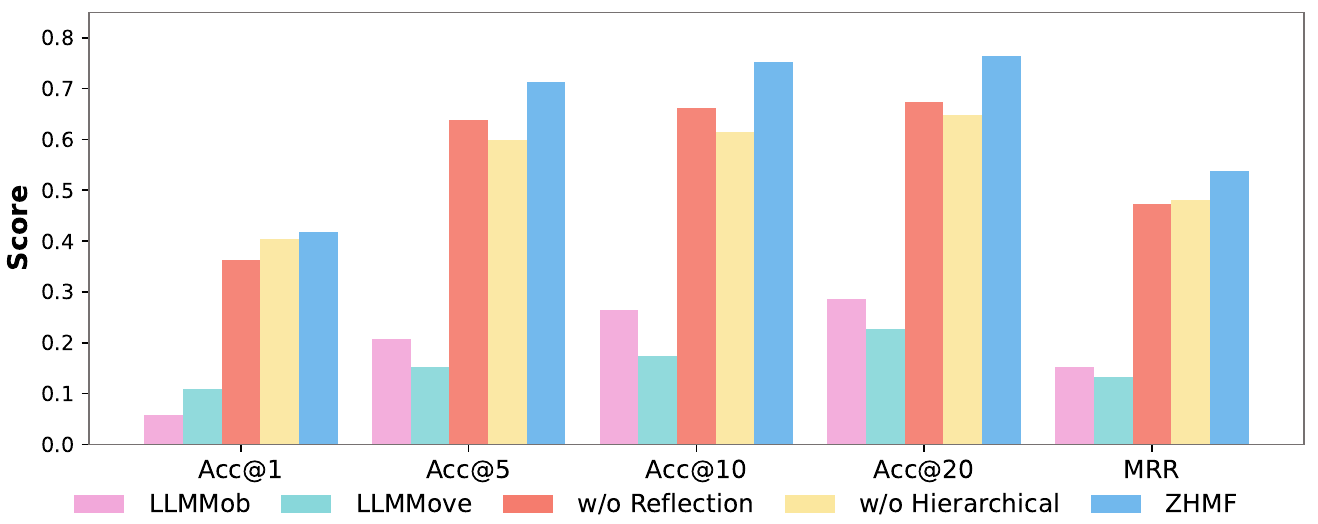}
    \caption{Zero-shot results on the NYC dataset.}
    \label{fig:zeroshot-bar}
\end{figure}

To evaluate our model's generalization for entirely unseen users, we adopt a zero-shot experimental setting: 30\% of users are randomly selected as "zero-shot users", with all their records excluded during training. The model is then evaluated solely on these users' test data, assessing its ability to recommend for new users absent during training. 
This setup evaluates whether our model can effectively transfer knowledge learned from existing users to new users who are not observed in the training phase, thus addressing the challenging zero-start scenario. 
Notably, only LLM-based methods can support the zero-shot setting, as other baselines depend on user embeddings learned from training users. 
Therefore, in the zero-shot evaluation, we only compared ZHMF with LLMMob and LLMMove. 
The results of this experiment on the NYC dataset are illustrated in Figure~\ref{fig:zeroshot-bar}.

As shown in Figure~\ref{fig:zeroshot-bar}, ZHMF achieves higher Acc@1 and MRR than both baselines in the zero-shot setting. These results indicate that our model can better generalize to users who were entirely absent from the training data and can still provide reasonable recommendations for such new users.

\subsection{Cold Start Forecasting}
To evaluate our model's ability to handle the cold start problem, which is a common challenge in recommendation systems, we conducted experiments to assess our model's performance across different user activity levels. 
We categorized users into three groups based on the number of trajectories in the training set: inactive users (bottom 30\% in trajectory count), normal users (middle 40\%), and very active users (top 30\% in trajectory count).  

\begin{figure*}[htbp]
    \centering
    \vspace{-3mm}
    \includegraphics[width=0.98\linewidth]{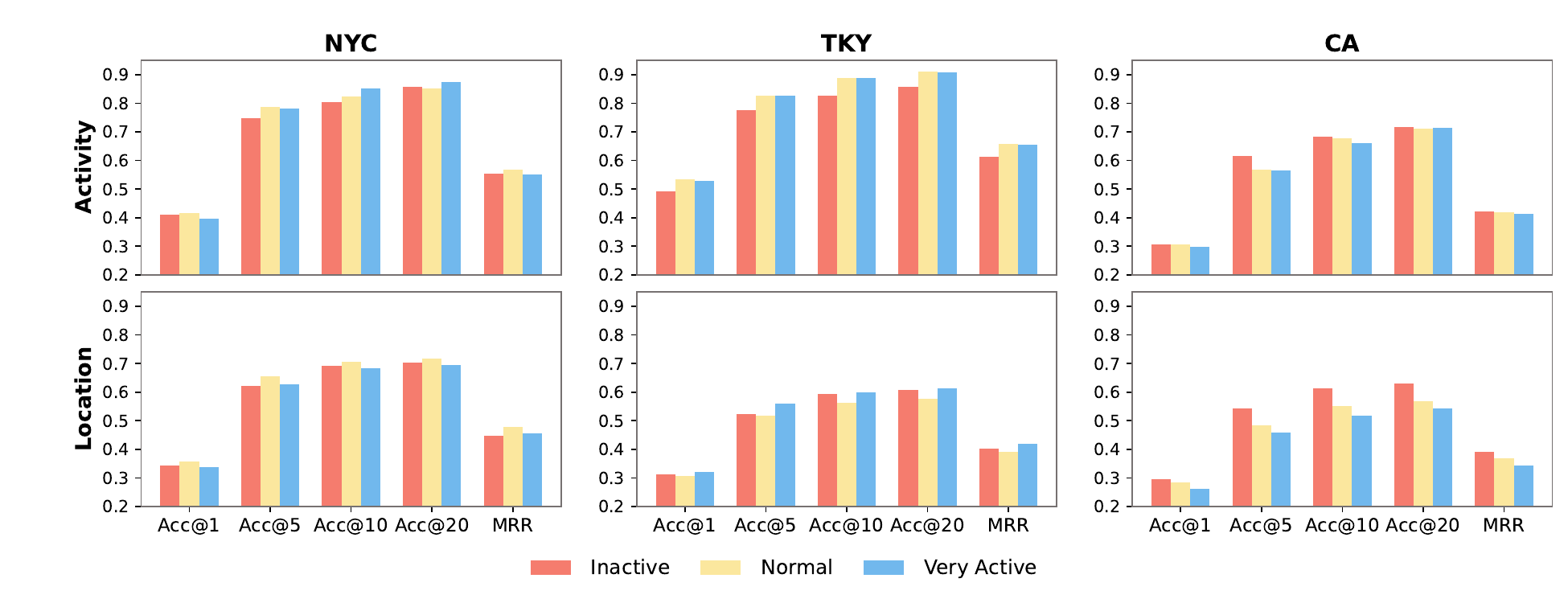} 
    \caption{Performance comparison under cold-start setting on three datasets.}
    \vspace{-3mm}
    \label{fig:cold_start}
\end{figure*}

This analysis aims to examine how effectively our hierarchical reasoning approach with reflective memory can leverage information from existing users to generate recommendations for users with limited historical data. Figure~\ref{fig:cold_start} visualizes the performance comparison across different user groups on three datasets in the cold-start scenario. Detailed results can be found in the supplementary materials. The results show that our model achieves strong performance for inactive users with limited histories, and the performance gap across activity levels is small. This indicates that our approach can effectively alleviate the cold start problem by leveraging transferable knowledge, especially for users with sparse trajectories.

\subsection{Ablation Study}




To assess the impact of each component in our model, we conducted ablation experiments on the NYC dataset with three model variants: 
ZHMF, the complete model that combines both the reflection mechanism and hierarchical reasoning; 
ZHMF-w/o-Reflection, which removes the reflection mechanism to evaluate its unique contribution while retaining the hierarchical reasoning process;
and ZHMF-w/o-Hierarchical, which removes the hierarchical structure to assess the effectiveness of the reflection mechanism alone. 
In each experiment, except for the replaced or removed components, all other settings remained consistent with the complete model. 
The experimental results are shown in Table~\ref{tab:ablation-nyc}. The complete model achieved the best performance, further validating the importance of both the reflection mechanism and hierarchical reasoning structure. 

\begin{table}[htbp]   
\centering  
\footnotesize
\caption{Ablation Studies of ZHMF on NYC Dataset}  
\label{tab:ablation-nyc}  
\setlength{\tabcolsep}{3pt}
\begin{tabular}{lccccc}  
\toprule  
Variants & Acc@1 & Acc@5 & Acc@10 & Acc@20 & MRR \\
\midrule  
w/o Reflection & 0.2866 & 0.6088 & 0.6607 & 0.6800 & 0.4151 \\
w/o Hierarchical & 0.3185 & 0.5523 & 0.6310 & \textbf{0.7075} & 0.4229 \\
ZHMF & \textbf{0.3437} & \textbf{0.6325} & \textbf{0.6904} & 0.7008 & \textbf{0.4596} \\
\bottomrule  
\end{tabular}  
\end{table}

Specifically, when comparing ZHMF with ZHMF-w/o-Reflection, the top-1 accuracy decreased from 34.37\% to 28.66\%, demonstrating the significant role of the reflection mechanism in enhancing recommendation performance. 
Similarly, the performance of the ZHMF-w/o-Hierarchical version was also lower than the complete model, with Acc@1 dropping to 31.85\%, suggesting that hierarchical reasoning is crucial for understanding user dynamic behaviors.
Furthermore, the hierarchical reasoning process and retrieval-augmented reflection enhance the interpretability of our framework, enabling users and developers to trace the key factors behind each recommendation.

\subsection{Hyperparameter Analysis}
To evaluate the sensitivity of the proposed ZHMF model to different hyperparameter settings, we conducted experiments on the NYC dataset. These studies aimed to explore how key hyperparameters impact model performance, optimize parameter configurations, and provide guidance for future research. In this experiment, we focused on two primary hyperparameters: the number of categories for activity-level decisions and the number of reflections in the reflection mechanism.

\noindent\textbf{Number of Categories for Activity-level Decisions.} The number of categories for activity-level decisions refers to the number of potential user target activity categories predicted by the activity-level policy module within the hierarchical reasoning framework. This parameter directly affects the model's ability to capture users' long-term preferences and the diversity of activity-level decisions. In this experiment, we tested the number of categories ranging from 5 to 30 while keeping other parameters unchanged.


\begin{table}[htbp]   
\centering  
\small
\setlength{\tabcolsep}{4pt}
\caption{Effect of activity-level decision categories on ZHMF performance (NYC dataset).}  
\label{tab:activity-level-categories}  
\begin{tabular}{cccccc}  
\toprule  
Categories & Acc@1 & Acc@5 & Acc@10 & Acc@20 & MRR \\
\midrule  
5   & 0.2999 & 0.6154 & 0.6711 & \textbf{0.7068} & 0.4303 \\
10  & 0.3014 & 0.6206 & 0.6667 & 0.6830 & 0.4293 \\
15  & 0.2769 & 0.5880 & 0.6362 & 0.6488 & 0.4025 \\
20  & \textbf{0.3437} & \textbf{0.6325} & \textbf{0.6904} & 0.7008 & \textbf{0.4596} \\
25  & 0.2977 & 0.6154 & 0.6615 & 0.6778 & 0.4250 \\
30  & 0.2888 & 0.6095 & 0.6585 & 0.6719 & 0.4181 \\
\bottomrule  
\end{tabular}  
\end{table}  

The results (see Table~\ref{tab:activity-level-categories}) demonstrate that as the number of categories increases, the model's performance first improves and then declines. When the number of categories is set to 20, the model achieves the best performance on metrics such as Acc@1, Acc@5, and MRR. A smaller number of categories provides moderate performance, while an excessive number of categories introduces noise, weakening the model's decision-making ability and leading to decreased accuracy.


\label{sec:hyperparameter}
\noindent\textbf{Number of Reflections in Reflection Mechanism.} 
The number of reflections refers to the number of historical reflection records retrieved by the model for each decision. This parameter determines the extent to which the system can utilize past experiences to enhance performance on the current task. In this experiment, we tested values ranging from 1 to 5 for the number of reflections.


\begin{figure}[htbp]
    \centering
    \includegraphics[width=1\linewidth]
    {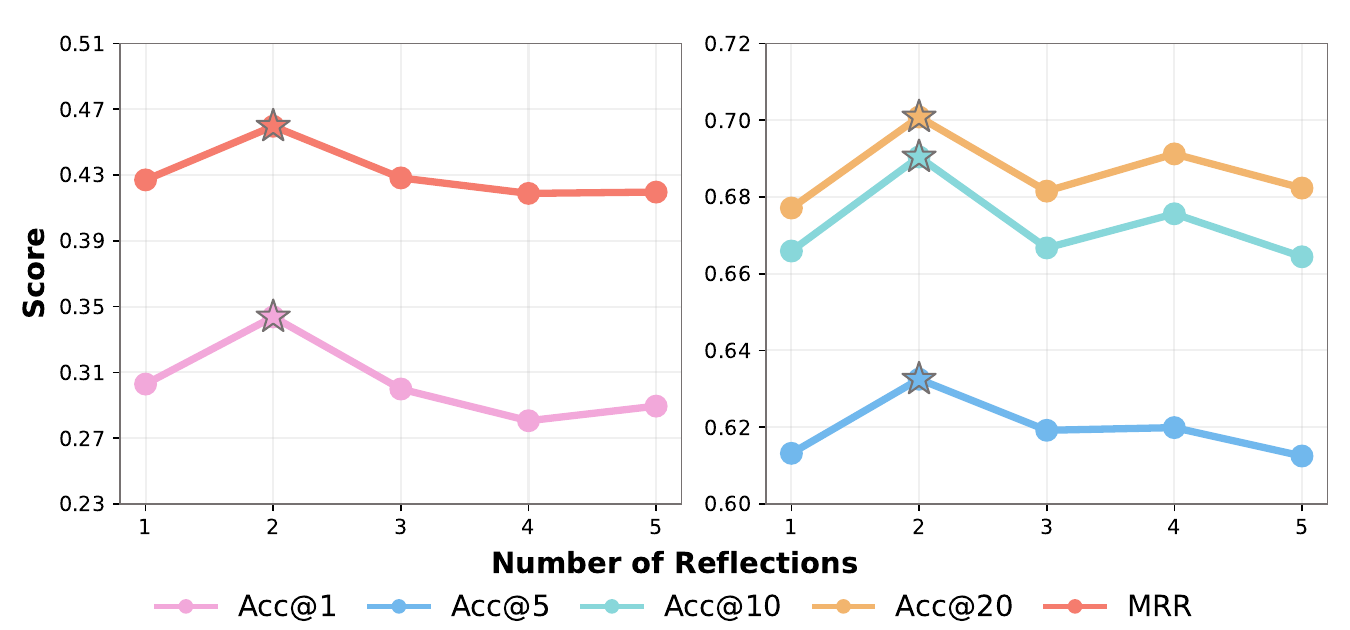}
    \caption{The impact of the number of reflections on the performance of ZHMF on the NYC dataset.}
    \label{fig:reflection-numbers}
\end{figure}

As shown in Figure~\ref{fig:reflection-numbers}, increasing the number of reflections from 1 to 2 improves the model's performance, and the best results are also obtained when the number of reflections is set to 2. As the number of reflections increases beyond 2, the performance does not further improve and even shows a slight decrease. This suggests that introducing too many historical records may bring redundant or less relevant information, which could affect decision efficiency.

\section{Conclusion}
\label{sec:conc}

\textbf{Summary.} 
We proposed ZHMF, a novel framework combining LLMs and hierarchical reasoning for human mobility modeling and location recommendation. 
By jointly modeling long-term patterns and short-term cues with a reflection-enhanced memory system, our approach delivers more accurate location intelligence, as shown by performance on three real-world datasets. 
ZHMF achieves substantial improvements under cold-start and zero-shot scenarios, demonstrating strong generalization and robustness in challenging real-world applications.

\noindent\textbf{Outlook.} Future work will improve spatial reasoning of LLMs, optimize memory retrieval efficiency, and enhance cross-domain generalization.
We also plan to integrate richer contextual modalities and consider replacing the reflection module with lightweight language models for greater efficiency and practicality. 
Enhancing accuracy and interpretability empowers intelligent travel services and smart city applications, improving user experience.

\noindent\textbf{Limitations.} Inherent uncertainties of LLMs, such as hallucinations and repetition, can undermine the quality and user trust. 
Significant variations in location categories across datasets challenge model generalization and limit cross-domain effectiveness.
This work focuses on single-step prediction and does not address multi-step forecasting or more complex recommendation, restricting real-world applicability.
Societal impacts of LLM-based mobility models, such as fairness and responsible data use, require further study.

\bibliography{reference}




\end{document}